# Cluster coloring of the Self-Organizing Map: An information visualization perspective

Peter Sarlin and Samuel Rönnqvist
Turku Centre for Computer Science – TUCS
Department of Information Technologies, Åbo Akademi University
Turku, Finland
{Peter.Sarlin, Samuel.Ronnqvist}@abo.fi

*Abstract*— **This paper takes an information visualization perspective to visual representations in the general SOM paradigm. This involves viewing SOM-based visualizations through the eyes of Bertin's and Tufte's theories on data graphics. The regular grid shape of the Self-Organizing Map (SOM), while being a virtue for linking visualizations to it, restricts representation of cluster structures. From the viewpoint of information visualization, this paper provides a general, yet simple, solution to projection-based coloring of the SOM that reveals structures. First, the proposed color space is easy to construct and customize to the purpose of use, while aiming at being perceptually correct and informative through two separable dimensions. Second, the coloring method is not dependent on any specific method of projection, but is rather modular to fit any objective function suitable for the task at hand. The cluster coloring is illustrated on two datasets: the iris data, and welfare and poverty indicators.**

## I. INTRODUCTION

The Self-organizing map (SOM), proposed by Kohonen [1], has been widely applied in fields like medicine, engineering and finance (see, e.g., [2,3]). It is an unsupervised neural network approach that pursues a simultaneous clustering and projection of high-dimensional data. Thus, the SOM can be thought of as a spatially constrained form of *k*-means clustering, or as a projection to a predefined and regularly shaped grid maintaining neighborhood relations in data. However, the regular grid shape, while holding merits for linking information to it, sets some restrictions. In particular, it leads to the need for additional visual aids to fully represent structures.

The SOM units may be seen as crisp clusters on the regular grid. Apart from being utilized as a stand-alone clustering technique, the SOM has also been used to serve input to a second-level clustering. Lampinen and Oja [4] proposed a two-level clustering by feeding the outputs of the first SOM into a second SOM. Further, Vesanto and Alhoniemi [5] illustrate that a two-level approach of the SOM, with hierarchical and *k*-means clustering, outperforms stand-alone techniques. Nevertheless, other approaches are needed if one aims at assessing distance structures on the grid. An early approach to information extraction is the U-matrix, where a color code between all neighboring nodes indicates their distance [6]. However, the U-matrix approach gives insights only into the local distance structures. With the aim of judging the degree of membership of a unit in a second-level cluster, fuzzy clustering and distance-based fuzzifications have been applied to the SOM [7,8]. Yet, fuzzifications only provide means to assess the distance structures with respect to individual clusters.

For the aim of illustrating global distance structures, the cluster coloring proposed by Kaski et al. [9] holds most promise. They propose an objective function based upon Multidimensional Scaling (MDS) with additional constraints: to meet the demands of preserving local neighborhood relations, and a color space with as much as possible of the variation occurring in two dimensions of hue, namely blue to yellow and red to green. However, while being an attractive approach, it comes with a number of limitations: *i*) the objective function takes a complex form and involves a number of parameters to be specified, *ii*) the coloring method is not flexible for different types of projection methods, and *iii*) the second constraint of Kaski et al., while aiding interpretation, distorts the color mapping. Recently, [10] introduced an approach to coloring that is appealing due to its simplicity. The colors of SOM units are derived through an approach that makes use of Principal Component Analysis (PCA) to derive three principal components of multivariate data and mixes them into RGB color channels with a linear function. Yet, their coloring lacks easily separable dimensions.

This paper starts by taking an information visualization perspective to the general SOM paradigm. We relate visual representations of the SOM to information visualization in general, with a key focus on describing the SOM from the viewpoint of data graphics. This involves viewing the SOM through the eyes of Bertin [11] and Tufte [12]. From this, we turn to the use of color spaces, and their combination with projection methods, to reveal cluster structures.

We aim at finding a general, yet simple, solution to cluster coloring. The approach follows that in Kaski et al. [9] by projecting multivariate SOM units into two dimensions and pairing the projection with a color space. The advantages of our approach relate to two aspects. First, it is central that the proposed color spaces should be easy to construct and customize to the purpose of use, while being perceptually correct and uniform. Second, the coloring method should not be dependent on any specific method of projection, but rather be modular to fit any objective function suitable for the task at hand.

The rest of the paper is organized as follows. First, we discuss perception, cognition and data graphics, as well as their relation to color spaces. Second, we introduce the SOM, continuous projection methods and the proposed color spaces. Then, we illustrate the proposed approach to cluster coloring through two experiments, before concluding the paper.

## II. PERCEPTION, COGNITION AND GRAPHICS

With the aim of pairing the SOM with continuous projection methods and color spaces to facilitate the interpretation of global distance structures, this section draws upon the information visualization literature. That is, we focus on three key aspects of information visualization: human perception, human cognition and data graphics. Along these lines, the key question is: *How to match the design of visuals according to the capabilities and limits of the human information and visual system?*

The visual system comprises the human eye and brain and can be seen as an efficient parallel processor with advanced pattern recognition capabilities (see, e.g., [13]). Mostly, arguments about the properties and perception capabilities of the human visual system rely on two grounds: *i*) information theory [14] and *ii*) early findings related to preattentive processing and gestalt theory. First, *information theory* states that the visual canal is best suited to carry information to the brain as it is the sense that has the largest bandwidth. Second, Ware [13] asserts that there are two main psychological theories for explaining how to use vision to perceive various features and shapes: *preattentive processing theory* by Triesman [15] and *gestalt theory* by Koffa [16]. Prior to focused attention, preattentive processing relates to simple visual features that can be perceived rapidly and accurately, and processed effectively at the low level of the visual system. Preattentive processing is useful in information visualization as it enables rapid dissemination of the most relevant visual queries. At a higher cognitive level, gestalt theory suggests that our brain and visual system follow a number of principles when attempting to interpret and comprehend visuals.

More related to the cognition of visualizations, Fekete et al. [17] relate the core benefit of visuals to them functioning as a frame of reference or temporary storage for human cognitive processes. The authors assert that visuals augment human memory, and thus enable allocating a larger working set for thinking and analysis. Along these lines, Card et al. [18] pinpoint how well-perceived visuals could amplify cognition: *i*) by increasing available memory and processing resources; *ii*) by reducing the search for information; *iii*) by enhancing the detection of patterns and enabling perceptual inference operations; *iv*) by enabling and aiding the use of perceptual attention mechanisms for monitoring; and *v*) by encoding the information in an interactive medium.

This clearly relates to the design of data graphics. The early, yet brilliant, work on principles for designing visual representations by Bertin [11] and Tufte [12] remain to be relevant. Their guidelines or rules of thumb, while being principles for graphics design, are also valid to overall computer-based visualizations. On the one hand, Tufte [12] puts forward a more focused set of principles for data graphics. His Theory of Data Graphics consists of two general principles: *i*) graphical excellence, and *ii*) graphical integrity. He defines graphical excellence as something that *"gives to the viewer the greatest number of ideas in the shortest time with the least ink in the smallest space"*. The second principle, graphical integrity, relates to telling the truth and avoiding misinterpretation about data. On the other hand, Bertin [11] puts forward a comprehensive framework called the Properties of the Graphic System, which consists of two planar and six retinal variables. The retinal variables are always positioned on the planar dimensions and can make use of three types of implantation: a point, line or area. In this paper, we mainly make use of Bertin's planar variables for positioning, and area implantations of the retinal variables of color and value (i.e., lightness or intensity). To this end, one key component relating to perception, cognition and data graphics is the notion of color spaces: *How should colors be derived to be perceptually correct, yet still differentiating and informative?*

Color spaces represent means of mapping colors to coordinates, which commonly refers to a three-dimensional space. The common RGB color space describes colors in terms of red, green, and blue light intensities that match the light sources of computer displays, as well as the photoreceptor cells of the human eye. However, this parameterization of color is arguably not the most intuitive to a viewer when focusing on relations among colors. Other models, such as HSL [19], commonly describe color by an independent lightness dimension, while additional dimensions describe properties such as hue (as a radial dimension) and saturation. Joblove and Greenberg [19] argue that humans primarily recognize hue, other properties secondarily, and that the HSL model therefore represents a more natural way of thinking about colors compared to RGB. A distinction is also present at the physiological level; colors are registered by cones in the human retina, whereas rods only register achromatic light intensity [20]. Thus, hue and lightness appear suitable for representing two dimensions so that they are easy to interpret separately.

The CIE 1976 ($L^*$, $a^*$, $b^*$) color space [21] (CIELAB) similarly uses an independent lightness dimension, whereas two orthogonal $a^*$ and $b^*$ dimensions jointly specify hue and saturation. The $a^*$ dimension positions a color between green and magenta and $b^*$ between blue and yellow. CIELAB attempts to provide an optimally uniform color representation, in which Euclidean distances in the color space should match perceived differences in color. We exploit the perceptual uniformity of CIELAB to correctly represent differences in data through color, thereby meeting Tufte's integrity principle.

As a central focus of this paper, we seek to define a color scale able to convey as much information as possible, while also being easily interpretable. In Tufte's terms, we seek graphical excellence, as well as concise and informative presentation. The color scale enables concise, preattentive communication of distances in data, and it is informative especially in that the viewer can intuitively distinguish the two underlying dimensions.

### III. SOMs, PROJECTIONS AND COLOR SPACES

This section describes the functioning of the SOM, MDS-based continuous projection methods and color spaces, as well as discusses how to combine them. Further, it also relates each subtopic to the above discussion on information visualization.

#### A. Information visualization with the SOM

The standard implementation of the batch SOM [1,2] based upon Euclidean distances is used herein. The SOM grid consists of a predefined number of nodes $m_i$ (where $i=1,2,…,M$), which represent reference vectors of the same dimensionality as the actual data set (i.e., number of inputs). Generally, the implemented batch SOM algorithm operates according to two steps:

1. Compare all data points $x_j$ with all nodes $m_i$ to find for each data point the nearest node $m_c$ in terms of the Euclidean distance (i.e., best-matching unit, BMU).
2. Update each node $m_i$ to averages of the attracted data, including data located in a specified neighborhood $\sigma$.

The neighborhood parameter $\sigma$ is herein specified automatically using the goodness measure by Kaski and Lagus [22]. The output of the SOM is a low-dimensional grid of units. As is common in the literature, and for the purposes herein, we use a two-dimensional grid of hexagonal units.

*So, how does the SOM relate to information visualization?* From the viewpoint of Bertin's [11] framework, the plane, and its two dimensions $(x,y)$, are described as the richest variables, which can be perceived at all levels of organization. On the SOM, they represent discrete neighborhood relations. This corresponds also to the key aim of the SOM, that is, to preserve neighborhood relations, whereas global distance structures are of secondary importance. The retinal variables, and their three types of implantation, are thus positioned on the grid. The six retinal variables may be used to represent properties of the SOM grid, particularly the units. They are as follows (where the parenthesis refers to Bertin's levels of organization): *size* (ordered, selective and quantitative), *value* (ordered and selective), *texture* (ordered, selective and associative), *color* (selective and associative), *orientation* (associative, and selective only in the cases of points and lines), and *shape* (associative).

The choice of retinal variable should be based upon the purpose of the visualization and the type of data to be displayed. For instance, variation in size has been used to represent frequency of data in units (see, e.g., [23]). Value, or brightness, has been used to visualize the spread of univariate variable values (i.e., feature planes) on the SOM (see, e.g., [23]). Likewise, texture has been used for representing cluster memberships (see, e.g., [24]). Orientation is commonly applied to represent high-dimensional reference vectors by the means of arrows (see, e.g., [2, p. 117]). Variation in color (or hue) has been used for illustrating crisp cluster memberships (see, e.g., [23]) and for a coloring that reveals cluster structures (see, e.g., [9]). Variation in shape is commonly used on the SOM by the means of labels, such as phoneme strings and phonemic symbols [2, pp. 208-210] (and the symbols in Figure 3b).

Relating to Tufte's [12] advise on graphical clarity and precision, the SOM can be said to link to multiple principles. As a projection from a high-dimensional space to one of a lower dimension obviously involves a loss of information, one approach to minimize distortions is to aim at only local (or otherwise limited) structure preservation, as the SOM does. This clearly relates to Tufte's advise on avoiding distortions of data. Furthermore, the regular, predefined grid shape of the SOM enables and facilitates many types of information linking to the same grid structure. This functions as an aid in thinking about the information rather than the design and encourages the eye to compare data. The SOM's property of approximating the probability density functions of data also facilitates presenting of vast data in a small space, as units will be located in dense areas of the data space, which could also be thought of as an aid in making large data sets coherent. On the SOM, data may be revealed at multiple levels of detail ranging from overview of multivariate structures on the grid, to illustration of individual data on the grid (e.g., trajectories located in their BMUs), which also integrates statistical and verbal descriptions. Along these lines, Tufte's six guidelines on telling the truth about data are also supported. For instance, showing data variation, not design variation, and not showing data out of context relates to, and is supported by, the use of a regular grid shape. Likewise, an example of visuals being directly proportional to the quantities they represent is the adjustment of color scales used for the linked visualizations, such as normalizations of feature planes in order for all variables to be comparable (see, e.g., [25]), and the use of perceptually uniform color scales, such as CIELAB.

Still, while the predefined and regular grid of the SOM is beneficial, it leads to the need for additional visual aids to fully represent structures in the multivariate SOM units.

#### B. Continuous projection methods

Following the approach in Kaski et al. [9], continuous projection methods can be used to reduce the dimensionality of the high-dimensional SOM reference vectors into a lower dimension. The first continuous projection methods date back to the early 20th century. However, only since the 1990's has there been a significant soar in the number of methods developed. We start by presenting one of the seminal projection methods, metric MDS [26]. Then, we relate Sammon's mapping [27] and Local MDS (LMDS) [28] to the functioning of standard MDS. The key aim of MDS-based methods is to project high-dimensional data $x_j$ to a two-dimensional data vector $y_j$ by preserving distances. Let the distance in the input space between $x_j$ and $x_h$ be denoted $d_x(j,h)$ and the distance in the output space between $y_j$ and $y_k$ be denoted $d_y(j,h)$. This gives us the objective function of metric MDS:

$$E_{MDS} = \sum_{k \neq l} \left( d_x(j,h) - d_y(j,h) \right)^2. \quad (1)$$

From the baseline MDS method, we can turn to counterparts that have an objective closer to that of the SOM: local distance preservation. Sammon's mapping [27] is an MDS method in that it attempts to preserve pairwise distances between data, but differs from standard MDS by focusing on local distances relative to larger ones. Likewise, LMDS [28] differs from metric MDS by restricting the objective function to local distances. The approach to turn the focus to pairs of points with small distances is borrowed from the force-directed graph layout algorithms, as the objective function introduces a repulsion between points with large distances.

The two-dimensional output of continuous projection methods illustrates global distance structures, which are represented by the planar dimensions (*x,y*) in Bertin's [11] framework. Along these lines, the key aim of the methods, while focusing on local distances, is on preserving absolute distances between data. However, we focus on turning these planar variables into retinal variables that can be represented on the SOM grid to illustrate structures.

*C. Color spaces*

While the regular grid structure of the SOM units restricts our use of Bertin's planar variables, different retinal variables can be used for illustrating distances on the SOM. We use value (lightness and saturation) and color (hue) in order to have variables that are ordered and selective. This can be approached through area implantations of colors from the CIELAB color space, as discussed in Section 2.

Based on the three-dimensional CIELAB space, we propose a straightforward approach for deriving two-dimensional subspaces (planes) to be used for visualization. Lightness (L*) is set as one of the dimensions in our two-dimensional scale, while the other dimension is specified as a line in the a*b* plane. The dimensions are intuitively separable to a viewer, as they vary in lightness and hue/saturation, respectively, and they are easier to interpret separately than when both dimensions vary in hue (see, e.g., [9]). Peripheral parts of the CIELAB space are not correctly displayable in RGB, and should be avoided to ensure perceptual uniformity in display. It is important to note that the safe area in hue (a* and b*) is irregularly shaped and narrower at more extreme values of lightness (L*).

We demonstrate our coloring approach by two different choices of plane, which are to be applied in SOM visualizations. The two white lines in Figure 1 illustrate the planes positioned in three cross sections of the CIELAB space: *i*) a green-yellow-red plane (L*=[20, 80], a*=[-60, 60], b*=40), and *ii*) a cyan-gray-red plane (L*=[20, 80], a*=[-45, 45], b*=a*). The colors of the two scales are presented in Figure 2. Our choices are not claimed optimal, but guided by a number of principles:

1. the extreme hue points should be perceived as clear-colored as possible in order to be easily identifiable as extremes
2. the scale should appear balanced, with an easily identifiable mid-point color
3. the number of differentiable shades should be maximized in both dimensions
4. the plane should not extend too far from the center, to avoid colors that are not correctly displayable in RGB.

In the first color scale, green and red are primary colors, thus deemed good for representing poles of the hue dimension; yellow serves as the mid reference point. The choice of b*=40 is a compromise between good color saturation (b* further from 0) and safe representation in RGB in those L* and a* intervals. The second scale is an example of using gray as the mid-point, where deviations are easily noticeable as fading towards cyan or red. The limit of displayable colors lies closer towards cyan than in other directions, which restricts the hue range also towards red, as gray is the fixed mid-point.

Since hue and lightness are perceived and thought of differently, it is challenging to compare distances between the dimensions. As Joblove and Greenberg [19] argue for hue being perceived primary to lightness, we find that hue plays a more dominant role than lightness in visually assessing cluster structures on the SOM. Therefore, we recommend to use the hue dimension to illustrate the first

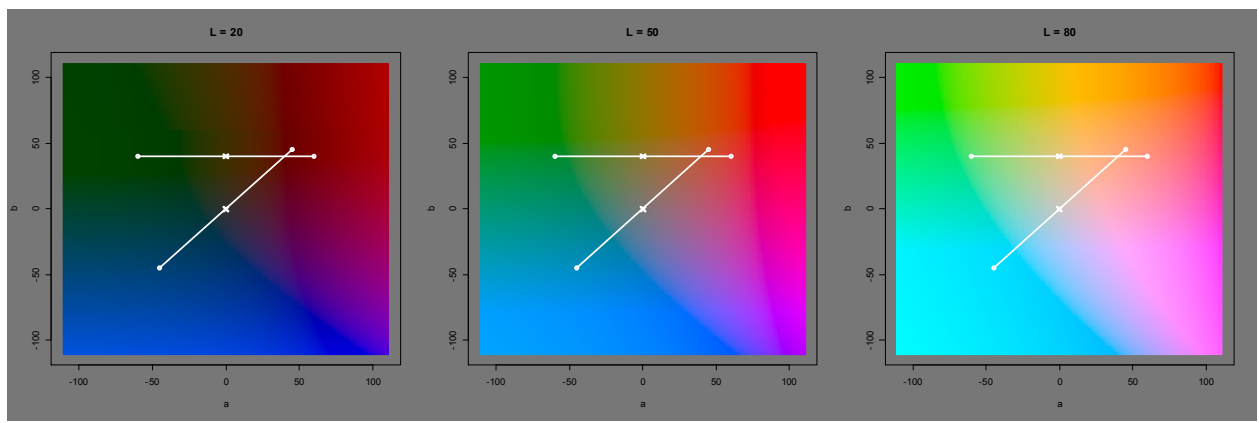

Figure 1. Two planes positioned in three cross sections of the CIELAB space, at extreme and mid points of lightness (L*).

principal component and lightness the second.

In our experiments in Section 4, we scale each of the two first principal components to the full range of our color scales. The rationale behind this is the emphasis on illustrating variation in data, at the risk of skewed ratios between components. This does, however, only invoke challenges in simultaneously perceiving distances in both dimensions.

Another complicating factor in using lightness, which is important to note, is that perceived lightness is dependent on context [20], namely the lightness of surrounding colors. We suggest for this reason that colors be presented with some sort of consistent reference color, to interleave colors representing data. The presentation in Figure 2 uses spacing between colors for this reason, to expose the reference background color, which is also applicable to presentation of the SOM.

## IV. EXPERIMENTS

This section illustrates the use of the cluster coloring of the SOM. We present experiments on two types of data: one UCI repository dataset and one real-world dataset. The iris [29] dataset functions as toy data in that it exhibits expected patterns, while we use World Development Indicators to illustrate the application of the coloring approach to a real-world case.

### A. Iris data

The iris dataset [29] comprises 150 instances of three classes (i.e., species) of iris flowers, where each instance is described by four variables. Prior to training a SOM, the data have been standardized to unit variance for equal weighting of the variables. We train a SOM with a 6x7 grid. The training procedure is automated and follows the description in Section 3.1. The SOM grid is presented in Figure 3b, which represents units with circles rather than hexagons to leave space for reference coloring. In this example, we use the cyan-gray-red color scale and a white reference color. We apply Sammon's mapping to the reference vectors of the SOM, and use its two principal components as an input to the color space. This produces the mapping in Figure 3a, and the coloring of the grid in Figure 3b. The grid is also overlaid with the iris data, differentiating classes using shapes. This illustrates that *setosa* (circles) are well-separated, whereas *versicolor* (triangles) and *virginica* (rectangles) are non-separable, as is also noted in early work [29]. The coloring of the SOM confirms this by illustrating larger distances between setosa and the two other classes, and only minor differences in distances between the overlapping classes of versicolor and virginica.

### B. Millennium Development Goals

The second application illustrates how the coloring can be applied to provide insight into real-world data, without prior information about classes. It uses a selection of World Development Indicators to illustrate the structures of welfare and poverty in the world. The World Development Indicators are collected by the World Bank and have commonly been used for demonstrating SOM processing and its extensions (see, e.g., [9,30-32]). The indicators measure the eight Millennium Development Goals (MDGs) that represent commitments to reduce poverty, ill-health, gender inequality, environmental degradation and the lack of education, and to improve access to clean water. The dataset consists of 15 indicators for 207 economies spanning from 1990–2008. Again, we standardize each indicator by variance for unit weighting of the SOM inputs.

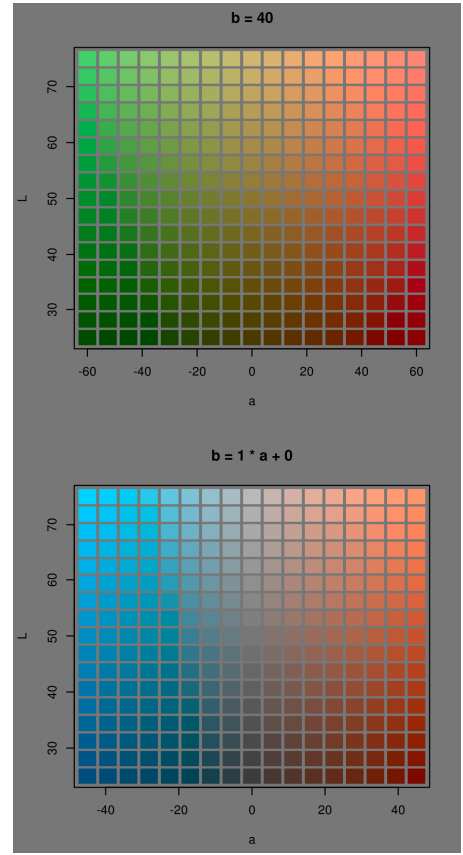

Figure 2. The two planes in CIELAB space, to be used for perceptually uniform SOM coloring.

We train a 9x9 SOM. In contrast to the earlier case, we use a color scale that differentiates in green to red hue and apply LMDS to the reference vectors of the SOM. The two-principal components of LMDS are used as an input to the color space, and thus to the coloring of Figures 3a and 3b. On top of the SOM grid, we have overlaid labels of an illustrative sample of economies in 2005. The labels illustrate, as was also shown in [33], that the upper left corner is populated mainly by Sub-Saharan economies, and the lower left corner by equatorial economies, mostly in Latin America. While the right part of the map mainly represents developed economies, the lower corner consists of Middle Eastern oil producing economies, and is characterized by high per capita pollution, whereas the upper part is characterized by high gender equality and less pollution, such as Northern European economies. The cluster coloring in Figure 3b illustrates large distances in the upper part between least developed economies (green) and developed economies (red, both dark and light). In comparison, the lower part shows clearly smaller distances, which are displayed in lighter colors. Thus, differences between the upper and lower part on the right side are smaller than differences between the upper and lower part on the left side.

## V. CONCLUSIONS

This paper views the general SOM paradigm from an information visualization perspective, focusing particularly on data graphics. We relate visualizations of the SOM to the ideas of Bertin and Tufte. Then, from the viewpoint of information visualization, we make use of coloring to reveal cluster structures.

The method for coloring the SOM aims at illustrating cluster structures by colors that represent distances in data. Projection methods are used to reduce the dimensionality of SOM units into two dimensions. This enables a data-driven coloring using a two-dimensional color scale, which communicates variation in the two separable dimensions of lightness and hue. The advantages of our approach relate to its simple, general and modular form. The proposed color spaces are easy to construct and can be combined with any continuous projection method.

The cluster coloring is illustrated on two datasets: the iris data and welfare and poverty indicators. This provides an illustration of expected patterns of the three classes of iris flowers and exploratory approach to assessing a real-world case.

Future work should focus on better integrating the fields of information visualization and dimension reduction. Particularly important topics are the integration of interaction techniques and user evaluations. We provide only a starting point to such a discussion, which hopefully inspires further interest and effort on the topic.

## ACKNOWLEDGMENT

The authors are grateful to Hanna Rönnqvist for insightful comments and discussions, and the Academy of Finland (grant no. 127592) for financial support.

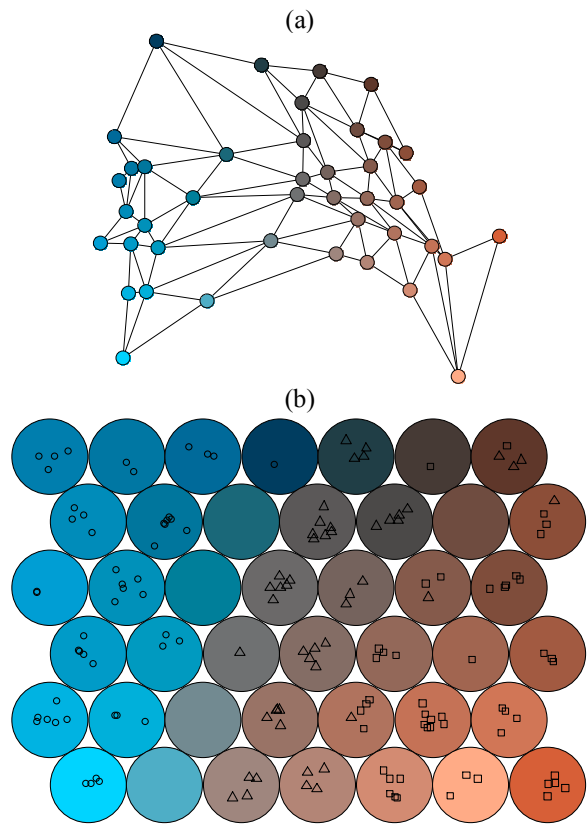

Figure 3. Iris data: (a) Sammon's mapping on the SOM units, and (b) derived coloring of the SOM.

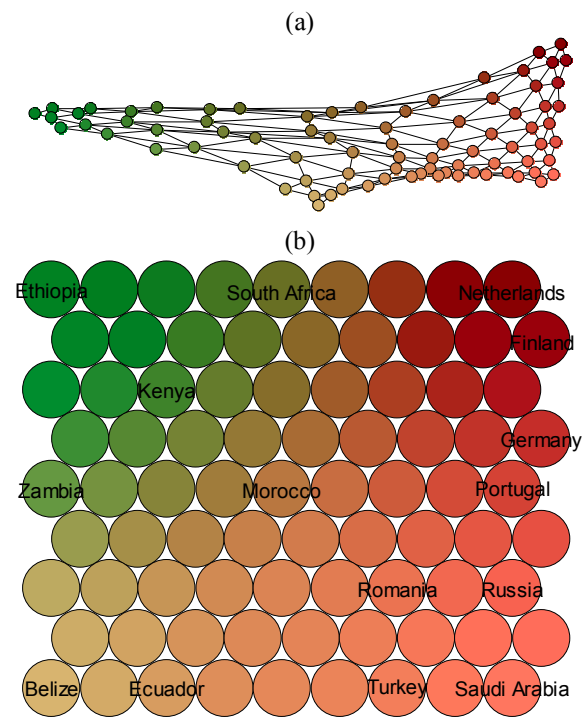

Figure 4. MDG data: (a) LMDS on the SOM units, and (b) derived coloring of the SOM.